\documentclass[conference]{IEEEtran}
\IEEEoverridecommandlockouts
\usepackage[inkscapelatex=false]{svg}
\usepackage{svg}
\usepackage{cite}
\usepackage{amsmath,amssymb,amsfonts}
\usepackage{algorithmic}
\usepackage{graphicx}
\usepackage{textcomp}
\usepackage{xcolor}
\usepackage{comment}
\usepackage{tabu}   
\usepackage{booktabs}      
\newcolumntype{L}[1]{>{\raggedright\let\newline\\\arraybackslash\hspace{0pt}}m{#1}}
\newcolumntype{C}[1]{>{\centering\let\newline\\\arraybackslash\hspace{0pt}}m{#1}}
\newcolumntype{R}[1]{>{\raggedleft\let\newline\\\arraybackslash\hspace{0pt}}m{#1}}
\def\BibTeX{{\rm B\kern-.05em{\sc i\kern-.025em b}\kern-.08em
    T\kern-.1667em\lower.7ex\hbox{E}\kern-.125emX}}
\usepackage{gensymb}

\usepackage{subcaption}
\usepackage{hyperref}
\usepackage{balance}

\newcommand{\tocite}[1]{\textcolor{red}{[CITE]}}
\newcommand{\todo}[1]{\textcolor{orange}{[TODO]}}
\newcommand{\tocheck}[1]{\textcolor{cyan}{[CHECK]}}

\begin{document}

\title{Assisting MoCap-Based Teleoperation of Robot Arm using Augmented Reality Visualisations\\
\thanks{This research is partially supported by the Australian Research Council Discovery Early Career Research Award (Grant No. DE210100858).}
}

\author{\IEEEauthorblockN{ Qiushi Zhou}
\IEEEauthorblockA{%\textit{Department of Computer Science} \\
\textit{Aarhus University}\\
%Aarhus, Denmark \\
0000-0001-6880-6594}
\and
\IEEEauthorblockN{Antony Chacon}
\IEEEauthorblockA{%\textit{School of Computing and Information Systems} \\
\textit{The University of Melbourne}\\
%Melbourne, Australia \\
0000-0001-5888-1646}
\and
\IEEEauthorblockN{ Jiahe Pan}
\IEEEauthorblockA{%\textit{Department of Mechanical and Process Engineering} \\
\textit{ETH Zurich}\\
%Zurich, Switzerland \\
0000-0003-2272-0925}
\and
\IEEEauthorblockN{ Wafa Johal}
\IEEEauthorblockA{%\textit{School of Computing and Information Systems} \\
\textit{The University of Melbourne}\\
%Melbourne, Australia \\
0000-0001-9118-0454}
}

\maketitle
%\vspace{-7em}
\begin{abstract}
Teleoperating a robot arm involves the human operator positioning the robot's end-effector or programming each joint. Whereas humans can control their own arms easily by integrating visual and proprioceptive feedback, it is challenging to control an external robot arm in the same way, due to its inconsistent orientation and appearance. We explore teleoperating a robot arm through motion-capture (MoCap) of the human operator's arm with the assistance of augmented reality (AR) visualisations. We investigate how AR helps teleoperation by visualising a virtual reference of the human arm alongside the robot arm to help users understand the movement mapping. We found that the AR overlay of a humanoid arm on the robot in the same orientation helped users learn the control. We discuss findings and future work on MoCap-based robot teleoperation.
\end{abstract}


\section{Introduction}

Robot arm is a popular type of robot for a wide range of general-purpose tasks thanks to their efficient form-factors. They are typically situated on desktops in a vertical orientation. A similar movement range of the human arm is the space on one side of the body (Figure \ref{fig:tethas13456}), perpendicular to the orientation of the movement range of the robot arm. This difference in \textbf{orientation} presents a challenge for Motion Capture (MoCap) based teleoperation for human operators to anticipate the movement of the robot. Further, the difference in \textbf{appearances} of the human and robot arms challenges operators to understand how the joint rotations are mapped. 

We explore how augmented reality (AR) can assist MoCap-based teleoperation of a robot arm by rendering a virtual arm as visual reference that mediates the inconsistencies between the human arm and the robot arm. In Study 1, we investigate how a virtual arm rendered in AR next to the robot arm could help improve user performance and experience of a target reaching task. We implemented three conditions of the AR arm that are either in a human-like or robotic appearance, and either in the same orientation with the robot or with the human arm. We concluded that the optimal configuration is a human-like arm overlaid on the physical robot in the same orientation with it to assist the understanding of the control mapping and to ensure easy visual access. In Study 2, we evaluated this AR arm using a posture matching task. We found that the AR arm helped reduce the perceived physical demand, effort, and frustration. Most participants found the AR arm more helpful for learning the control at the beginning than as an always-on visual guidance for teleoperation.% Finally, we summarise findings that inform future work exploring MoCap teleoperation of robot arms. %We also describe a system implementation that integrates ROS, MoCap, and AR components for future reference.
\section{Background}

\subsection{Robot Teleoperation via Human Motion Mapping}

Robot teleoperation enables human operators to remotely control a robot to perform tasks that are difficult for the human body through manipulation interfaces~\cite{pepper1984research}. It requires extensive training, especially for robots with high DOF that challenge operators to anticipate the robot movement~\cite{Hedayati2018Improving}. Anthropomorphic robots afford the unique possibility to perform human-like movements for tasks that demand anthropomorphic appearances of movements, inspiring works such as~\cite{gulletta2021human} that generates human-like movement for human-like robot arms. While previous work found that anthropomorphism help users perceive robot actions~\cite{Hegel08Understanding,zlotowski15anthropomorphism,Roesler21meta}, no previous work has explored robot teleoperation control that capitalise on the ease of understanding anthropomorphic robotic movements. %While robot arms can move like human arms, it is plausible to control them using naturalistic movements of human arms, to enable potentially intuitive and ease-to-use manipulation interfaces for teleoperation systems. 
%We explore MoCap-based teleoperation of a robot arm through a 1:1 mapping of joint rotations with a human arm.

\subsection{AR Visualisations and Robot Control}

AR has been used for HRI for different purposes \cite{walker2023virtual}, such as to support real-time control and teleoperation (for reviews, see~\cite{Makhataeva20Augmented,Suzuki22Augmented}). For remote teleoperation, AR visualisations enhance situational awareness and reduce cognitive load by immersing the operator in a representation of the remote environment and overlaying information related to the task~\cite{yew2017immersive}. For co-located HRI tasks, AR can be used to visualise motion intent of robots intuitively by directly showing their potential movements within the physical space~\cite{rosen2019communicating}, allowing users to understand the motion intent more easily~\cite{Gruenefeld20Mind}. Whereas previous work explored AR-enabled robot control, such as using virtual shadows on physical floors to position drones~\cite{Chen2021PinpointFly} and interactive robot programming with virtual movement cues and anchors rendered in the physical environment~\cite{OSTANIN18Interactive}, it remains to be investigated how AR can help teleoperation by visualising interactive anthropomorphic movement cues for operators to learn the control using their own body movement. We explore how a virtual arm next to or overlaid on a physical robot arm can help with MoCap-based teleoperation.

\section{MoCap-Based AR Teleoperation System}
We explore how a virtual arm visualised in AR can help users learn the control while mitigating inconsistencies between the robot and the human arms regarding their \textit{Orientations} and \textit{Appearances} in Study 1, and determine the optimal configuration of the AR arm. In Study 2, we explore how it affects user performance and experience of a posture-matching task. Both studies were approved by the IRB of our university.

%\subsection{System Structure}

We utilised three key hardware components: 1) a 7-DOF robotic arm Franka Research 3 (FR3), 2) HoloLens 2, and 3) OptiTrack motion capture system to enable teleoperation of the FR3 by capturing the operator’s right arm movements. The OptiTrack 8-camera system captures the arm's orientation, while HoloLens 2 renders a virtual arm that corresponds to the movement of the physical robot arm in AR.

\begin{figure}
    \centering
  \includegraphics[width=.7\linewidth]{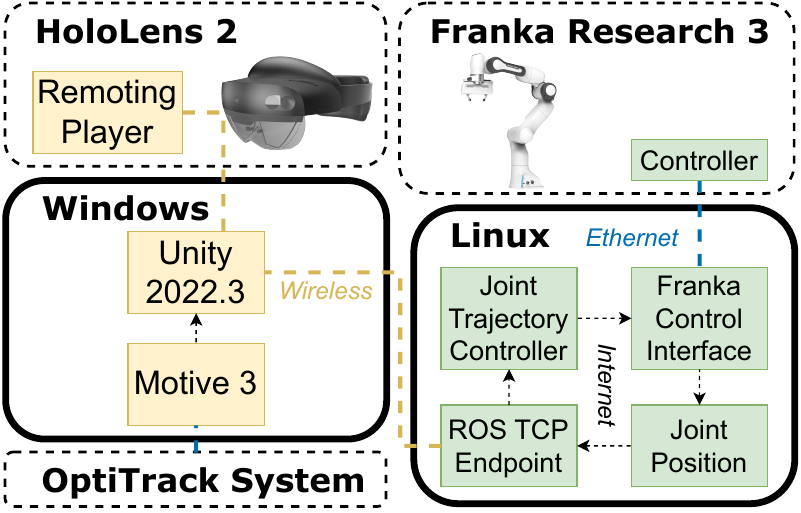}
  \caption{System architecture: a Linux machine running ROS 2 and the Franka Control Interface, a Windows machine running Unity and Motive, a Franka Research 3 robot, and a HoloLens 2 that renders through a remoting player.}
  \vspace{-1.5em}
  \label{fig:system}
\end{figure}

Robot control is handled by a Linux subsystem, which processes the motion commands through a custom joint trajectory controller and communicates with the Franka Control Interface. A ROS TCP Endpoint facilitates real-time data transfer between Unity and the robot controller, while joint position tracking monitors the robot's kinematics. Figure \ref{fig:system} illustrates the communication flow between these components.

\begin{figure}
    \centering
  \includegraphics[width=0.7\linewidth]{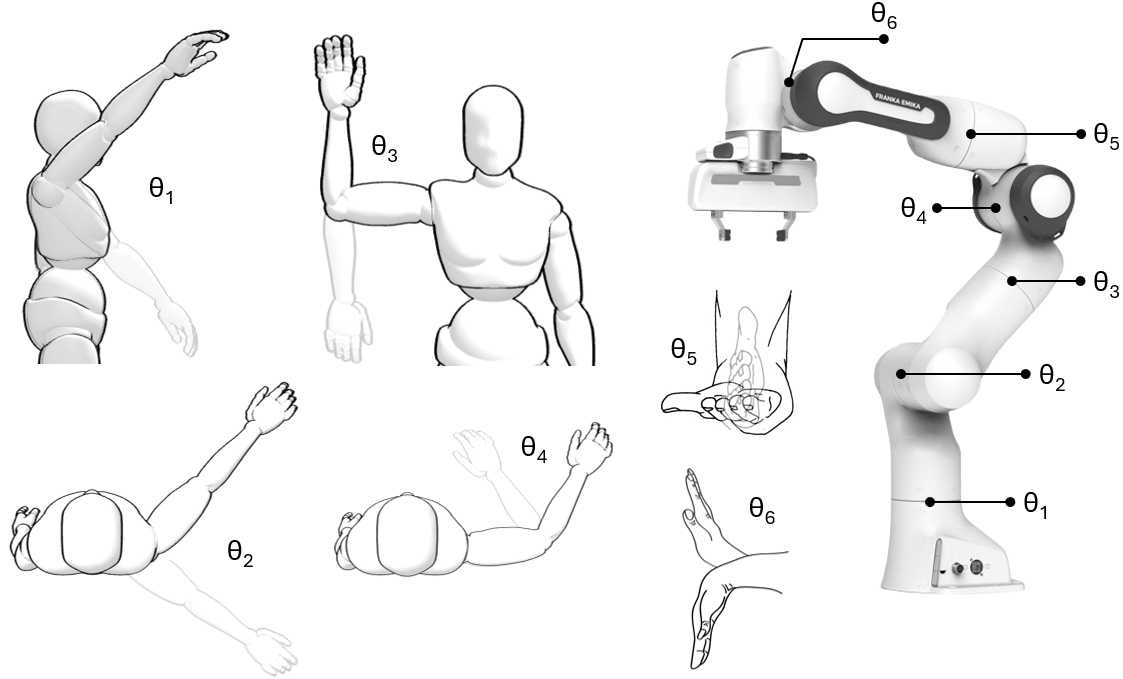}
  \caption{Corresponding joint angles of human and robot arms.}
  \vspace{-1em}
  \label{fig:tethas13456}
\end{figure}

%\subsection{Human to Robot Mapping}
%Model number, DOFs, reachable angles, etc..
We employed the FR3 robot because it resembles the structure of a human arm while maintaining the typical layout of an industrial 7-DOF manipulator. It features groups of joints that correspond to the shoulder (which rotates in three degrees of freedom), elbow (flexion/extension and forearm supination/pronation) and hand (flexion/extension). % Finally, although joint 7 is more suited for hand supination/pronation rather than directly replicating hand ulnar/radial deviation, we set its angle as the relative orientation of the back of the hand to the forearm, influenced by thumb and index finger movements. While the robot has some joints with non-zero offsets, this does not impact the mapping process, as we rely solely on the orientation of the arm segments. 
Adapting this mapping to other manipulators with a similar configuration would be straightforward. %Details about the calculations used to map human arm orientations to the robot’s joints are provided next.
The robot's joint angles are denoted as \(\theta_1\) through \(\theta_7\). %with the numbering starting from the robot's base.
The tracked positions of the shoulder, elbow, and wrist in three-dimensional space are represented as vectors: \(\mathbf{v}_{\text{shoulder}}\), \(\mathbf{v}_{\text{elbow}}\), and \(\mathbf{v}_{\text{wrist}}\), respectively. Each vector is expressed in the form \(\mathbf{v}_i = (x_i, y_i, z_i)\). Figures %\ref{fig:tetha2} and 
\ref{fig:tethas13456} illustrates the relationship between human arm motion and the robot's joint angles in a left-handed coordinate. Given the vector
\begin{equation}
\mathbf{v}_{\text{upper\_arm}} = \mathbf{v}_{\text{elbow}} - \mathbf{v}_{\text{shoulder}}
\label{eq:v_upperarm}
\end{equation}

and the robot orientation, we define the second joint \(\theta_2\) as
\begin{equation}
\theta_2 = \text{atan2}(z_{\text{upper\_arm}}, x_{\text{upper\_arm}})
\label{eq:theta2}
\end{equation}

The first joint \(\theta_1\) maps the elevation of the upper arm during flexion/extension. Thus, we determine it as
\begin{equation}
\theta_1 = \text{atan2}\left(y_{\text{upper\_arm}}, \sqrt{x_{\text{upper\_arm}}^2 + z_{\text{upper\_arm}}^2}\right)
\label{eq:theta1}
\end{equation}

%We obtain the third joint angle \(\theta_3\) within Unity's environment by extracting the local rotation of the upper arm object along the axis \(\mathbf{v}_{\text{upper\_arm}}\).
With 
\begin{equation}
\mathbf{v}_{\text{forearm}} = \mathbf{v}_{\text{wrist}} - \mathbf{v}_{\text{elbow}}
\label{eq:v_forearm}
\end{equation}
we determine the fourth joint angle \(\theta_4\) as
\begin{equation}
\theta_4 = \arccos\left(-\frac{\mathbf{v}_{\text{upper\_arm}} \cdot \mathbf{v}_{\text{forearm}}}{|\mathbf{v}_{\text{upper\_arm}}| \cdot |\mathbf{v}_{\text{forearm}}|} \right)
\end{equation}

The third and fifth joint angles, \(\theta_3\) and \(\theta_5\), map internal/external rotation of the upper arm and supination/pronation of the forearm respectively. Thus, we used the local rotation of these objects along the axes \(\mathbf{v}_{\text{upper\_arm}}\) and \(\mathbf{v}_{\text{forearm}}\) in Unity. Similarly, we derived the final two joint angles using the relative orientations obtained directly from Unity.

%The last two joint angles partially map wrist movements. \(\theta_6\) directly corresponds to hand extension/flexion (Figure~\ref{fig:tethas13456}), while \(\theta_7\) is more suitable for supination/pronation, rather than ulnar/radial deviation. \(\theta_7\) is determined by the relative angle of the back of hand generated by thumb and index finger movements. Likewise, both joint values are obtained in Unity.

%\begin{figure}
%    \centering
%  \includegraphics[width=0.6\linewidth]{figures/theta2.png}
%  \caption{Robot orientation comparing with human arm.}
%  \vspace{-1em}
%  \label{fig:tetha2}
%\end{figure}

%\begin{figure}
%    \centering
%  \includegraphics[width=\linewidth]{figures/theta6.png}
%  \caption{Sixth robot joint mapping wrist flexion/extension}
%  \label{fig:tetha6}
%\end{figure}

%\subsection{MoCap implementation}
%How movement tracking is implemented to work with the robot and with the AR
We employed an OptiTrack system and its software interface, Motive 3. Eight Prime-13W cameras were mounted on a tracking rig attached to the ceiling of the usability lab, spaced out and focused on the participants' right arm. We employed 3 sets of markers to track the orientation of the upper arm, forearm, and hand. Motive 3 transmits these data to Unity at an average speed of 200 KB/s. We used the HoloLens 2 and the Holographic Remoting Player app connected to a PC. The Mixed Reality Toolkit 3 (MRTK3) allowed us to control the HoloLens 2 in real-time through our Unity implementation. For calibration, we placed a QR code on the back of the robot at a height of 18 cm and 6 cm behind the center of its base. We used the Reality Collective package (https://github.com/realitycollective) to recognise the QR code via the HoloLens 2 cameras and to log its position. %Before starting the tasks, we instructed participants to face the QR code for the HoloLens to recognise the robot's position.
%\subsection{AR implementation}
%How AR was implemented with position tracking of robot, different visualisations, where they are anchored and how they are rotated, etc.. 

%Specify the 4 different conditions here (perhaps abbreviations?)
%\begin{itemize}
%    \item \textsc{Baseline}
%    \item \textsc{HumanHorizontal}
%    \item \textsc{HumanVertical}
%    \item \textsc{RobotHorizontal}
%\end{itemize}

%\begin{figure}
%    \centering
%  \includegraphics[width=.6\linewidth]{figures/mocap_markers.png}
%  \caption{MoCap markers placed on the arm. Markers for the upper arm and forearm were attached with kinesiology tape. We use the Quantum Metaglove's markers for wrist tracking.}
%  \label{fig:markers}
%\end{figure}
\section{Study 1: configuring the AR visualisation}

\begin{figure}
    \centering
    \includegraphics[width=\linewidth]{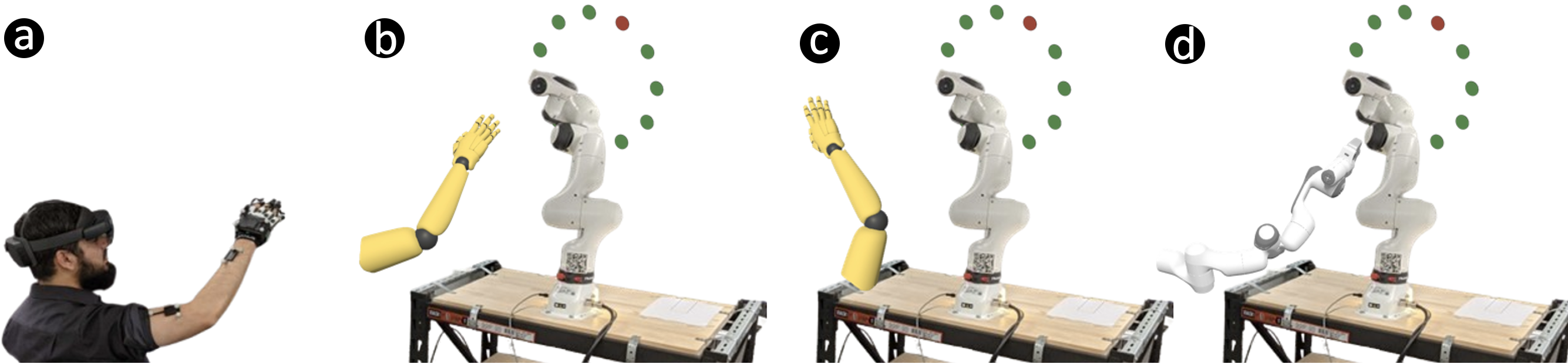}
    \caption{AR \textsc{Visualisation} conditions and apparatus: (a) Participant situated away from the robot; (b) \textsc{HumanHorizontal} arm in AR next to the physical robot; (c) \textsc{HumanVertical}; (d) \textsc{RobotHorizontal}.}
    \vspace{-1em}
    \label{fig:study1_AR_conditions}
\end{figure}

When a typical 7-DOF robotic arm (e.g., Franka Research 3, Kinova Gen3, etc.) is situated on a horizontal platform, its range of movement is similar to a human arm on the side of the body (Figure \ref{eq:theta2}). A challenge of teleoperating these robots is the mental effort for the human operator to mentally convert the rotation direction of their arm joints to those of the robot. Further, the different appearances between the human and the robot arm challenge the user to understand how the robot arm would respond to their control. We employ a within-subject design with four \textsc{Visualisation} conditions (Figure \ref{fig:study1_AR_conditions}) to compare and determine the optimal configuration: \textsc{HumanHorizontal (HH)}: A virtual human-like arm in the same orientation as the human arm. \textsc{HumanVertical (HV)}: A virtual human-like arm in the same orientation as the physical robot arm. \textsc{RobotHorizontal (RH)}: Virtual replica of robot arm in the same orientation as the human arm. \textsc{Baseline}: No AR visualisation rendered (Figure \ref{fig:study1_AR_conditions}).

%Using these conditions, we investigate how AR visualisations help users understand the control mapping between their arm joints and those of the robot to perform teleoperation tasks by providing a visual reference of the orientation (\textsc{RH}), appearance (\textsc{HV}), or both features (\textsc{HH}) of the human arm. 

%\subsection{Physical Setup and Task}
%How the robot is placed in the room of XxX m size, and where AR visualisation is placed relative to robot, and where participants are placed and why
The studies were conducted in a room with dimensions of 5.5$\times$3.6 m. We set the AR visualizations for HH, HV, and RH with respective offsets of (-0.8, 0.33, 0.4) m, (-0.5, 0.33, 0.4) m, and (-1, 0.33, 0.4) m relative to the robot. The participant sat 1 m to the left and 1.2 m behind the robot's base. %These distances were chosen to ensure comfortable visual access for the virtual arms within the field of view of the HoloLens 2. The participant was not positioned directly behind the robot to avoid occlusion of the virtual targets by the robot arm.

%\subsection{Task}

The task was reaching for a ring of 11 virtual circular targets with the robot's end-effector in a pre-defined sequence, as illustrated in Figure \ref{fig:study1_AR_conditions} (e)\cite{suarez2020benchmarks,funk2021benchmarking,Yu2020Engaging, Zhou2024reflected}. The ring of targets was parameterised by the radius \(R = 22.5\) cm and the diameter \(d = 5\) cm. The ring was positioned on a vertical plane directly in front of the robot. The centre of the ring was located at a displacement of (0, 0.56, 0.9) m relative to the robot's base, which was determined through pilot testing to allow the user's arm to move within a comfortable range. %The target reaching task was chosen as it has been employed to benchmark robot efficiency in manipulation tasks \cite{suarez2020benchmarks, funk2021benchmarking} and to evaluate interface design and human movement performance in HCI~\cite{wagner2023fitts, Yu2020Engaging, Zhou2024reflected}. 
The active target (red) turns green upon a successful selection, which requires the robot's end-effector position to be within a perpendicular distance of 10 cm from the target plane. Participants were asked to complete the task as fast as possible.

%\subsection{Participants and Procedure}
We recruited 24 participants (11 F, 13 M) with a mean age of 26.9 years ($Min=19, Max=36, SD=4.5$) using the University's online billboard. Participants rated their prior experience with VR and AR on a 7-point scale from 1 (never used) to 7 (use frequently) with a mean rating of 2.13 ($Min=1, Max=6, SD=1.45$). %Each participant took approximately 1.5 hours to finish the experiment and was compensated with a \$20 voucher. 
We calibrated HoloLens' built-in eye-tracker for each participant, and placed OptiTrack markers on their right arms. We used a 2-minute training block for participants to practice teleoperating the robot under the \textsc{Baseline}. Then, participants completed four rounds (two trials per condition) of tasks with breaks between trials. The condition orders were counterbalanced using a Latin square. After each condition, participants filled out a questionnaire on their perception of the task and system. After all tasks, participants ranked their preferences for the conditions. %The study took approximately 90 min.

%\subsection{Measures}
\subsection{Results}
Performance was evaluated through \textit{Movement Time}, the time interval between the appearance of the target and the successful selection. %For each trial, 11 timestamps of successful selection were recorded, from which 10 intervals were computed as the \textit{Movement Time} data points. This therefore resulted in: 10 \textit{time intervals} $\times$ 2 \textit{trials} $\times$ 4 \textit{\textsc{Visualisation}s} = 80 data points per participant. 
% where success requires the robot's end-effector position to be inside the circular target, and its perpendicular distance from the target plane to be within \todo{}.
% During each trial, we recorded the robot position information at \todo{}\,Hz. 
We administered a NASA-TLX questionnaire \cite{NASA-TLX} at the end of each condition. After finishing all tasks, participants ranked the four conditions based on their preferences, and answered interview questions: \textit{\textbf{Q1}. Do you think the AR arms were helpful for your controlling of the robot arm? \textbf{Q2}. Do you think it was more helpful with the virtual arm in the same orientation with the physical robot or with your own arm? \textbf{Q3}. Do you think it was more helpful seeing a virtual human arm or seeing a virtual robot arm?}

While we did not find statistically significant results in Movement Time or in NASA-TLX, user rankings of different \textsc{Visualisation} conditions showed that \textsc{HV} was the most preferred ($M=2.08, SD=.97$), followed by \textsc{HH} ($M=2.38, SD=.97$), \textsc{Baseline} ($M=2.54, SD=1.35$), and finally \textsc{RH} ($M=3.00, SD=1.02$).

%The post-study questionnaire yielded 90 responses. Two researchers collaboratively carried out a general inductive analysis, using independent parallel coding to categorise notable factors mentioned by participants that corresponded with their layout creation and adaptation style~\cite{Thomas2006general}. We present categories of the answers from participants in Table \ref{table:S1-qual} with the number of participants with answers in each category. 

%\subsubsection{Q1: Was AR helpful for the task?}

Twelve participants reported that the AR visualisation was helpful because it provided visual reference for them to conveniently see how the robot moves in correspondence to their control, without needing to to look back-and-forth between the robot and their arms: ``\textit{It's very helpful especially the vertical human arm ... I don't need to think of the direction (where) my arm goes to control the robot}(R3).'' Participants also mentioned that the virtual arm gave them confidence (P22), especially in how far the robot moves into distance (P15). Five reported that their attention was drawn to the targets and the robot instead of using the visualisation. %Four participants mentioned that the AR arm would be helpful for learning how the movement directions are mapped between the human arm joints and the robot arm joints at the beginning of the teleoperation, however, they may not be necessary after the user masters the control: ``\textit{As I got used to the movements, (the visualisation) could probably be ignored} (P19).'' ``\textit{(They were) sort of helpful. But as I was quite fast getting used to (the task) so I gradually reduced the frequency I referred to the AR arms} (P21).'' %Three participants did not provide specific reasons why they did not think the AR visualisations were helpful.

%\subsubsection{Q2: Which orientation was more helpful?}

Thirteen participants reported that the vertical orientation of the AR arm was more helpful because it was consistent with the physical robot, making it easier to understand the control. Seven participants preferred the horizontal orientation as it represents how their own arm behaves. Fifteen participants thought the \textsc{Human} arm was more helpful because it resembled their own arm close to the robot for visual reference. Nine of them deemed \textsc{HumanVertical} most helpful.

\subsection{Discussion }

While \textsc{Visualisation} did not yield statistically significant differences in movement time, qualitative feedback suggested that participants still found the virtual arm helpful. While the vertical orientation and the human-like appearance of the virtual arm were most preferred, it is also the only configuration of AR that can be overlaid on the physical robot for easier visual access. We choose \textsc{HV} as the optimal configuration.

\section{Study 2: Posture Control}

In Study 2, we investigate how an AR overlay of a human-like arm on the physical robot arm can assist MoCap-based teleoperation. We employ a posture matching task to avoid participants adapting to the repetitive movements as in target reaching, while requiring them to understand the mapping between their own arms and the robot. We employ a within-subject design that compares task performance and subjective measures between two \textsc{Visualisation} conditions: \textsc{AR Arm} where the virtual arm is presented during the tasks, and \textsc{No Arm} as a baseline without virtual arm. 

%\subsection{Task}

\begin{figure}
    \centering
    \includegraphics[width=.7\linewidth]{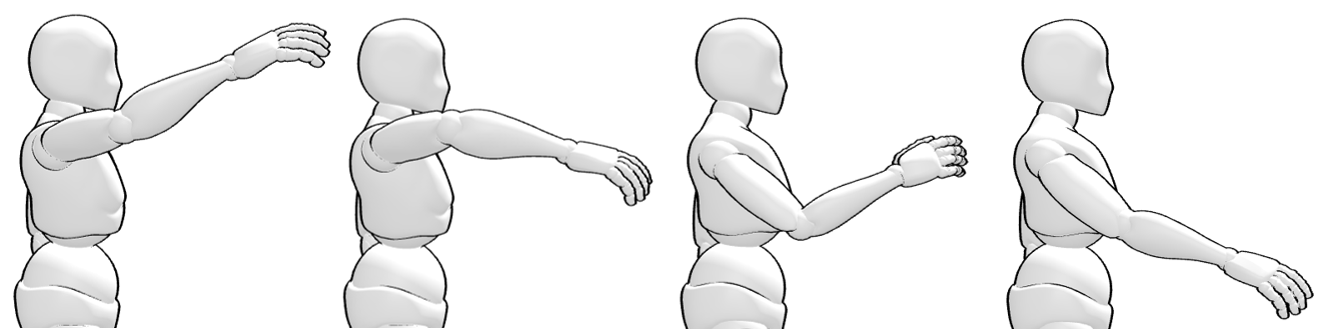}
    \caption{S2 postures: a) Elbow up, wrist up; b) Elbow up, wrist down; c) Elbow down, wrist up; d) Elbow down, wrist down.}
    \vspace{-1.5em}
    \label{fig:study2_postures}
\end{figure}

\begin{figure}
    \centering
    \includegraphics[width=.7\linewidth]{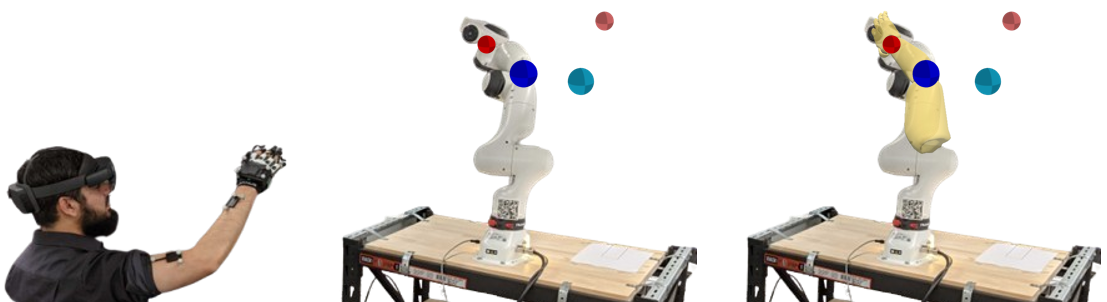}
    \caption{Study 2: participant, condition \textsc{No Arm} and \textsc{AR Arm}. Blue and red sphere rendered on elbow and wrist respectively. Lighter blue and red denote targets to match posture.}
    \vspace{-1.5em}
    \label{fig:study2}
\end{figure}

The task was sequentially matching postures using the real robot arm (Figure~\ref{fig:study2}). We rendered a blue sphere on the robot elbow and a red sphere on the wrist for visual reference. A target posture is matched when the elbow and wrist of the robot are both within 5 cm from their respective target positions (light blue and red). We selected 4 target postures (Figure~\ref{fig:study2_postures}) within a comfortable range of movement. The task was designed to model real-world scenarios where joint-space control of the robot is required to avoid collisions during teleoperation, which may arise in highly cluttered workspaces.

%\subsection{Participants and Procedure}
We recruited 24 participants (14F, 10M) with a mean age of 25.6 years ($SD=4.6$), mean rating of prior experience with VR and AR (1-7) of 2.04 ($SD=1.27$). Participants completed 2 rounds of the task under each condition. Each trial always started with the robot arm in a ``straight" posture with full extension. The order of postures were counterbalanced using a Latin square. Participants filled out a questionnaire after each condition, and received an interview after all conditions for feedback on their perception of the task and visualisation. %The study took approximately 50 minutes.
\vspace{-0.2em}
\subsection{Results}

\textit{Movement Time} is defined as the time taken to successfully match each target posture from when they first appear. We administered NASA-TLX \cite{NASA-TLX} in the same manner as in Study 1. In the end, participants answered the following interview questions: \textit{\textbf{Q1}. Do you think the AR arms were helpful for your controlling of the robot arm? \textbf{Q2}. Have you learned how to control the robot arm from seeing the virtual arm?} 

\begin{figure}[t]
\centering
\begin{subfigure}[t]{0.35\linewidth}
    \centering
    \includegraphics[width=0.95\textwidth]{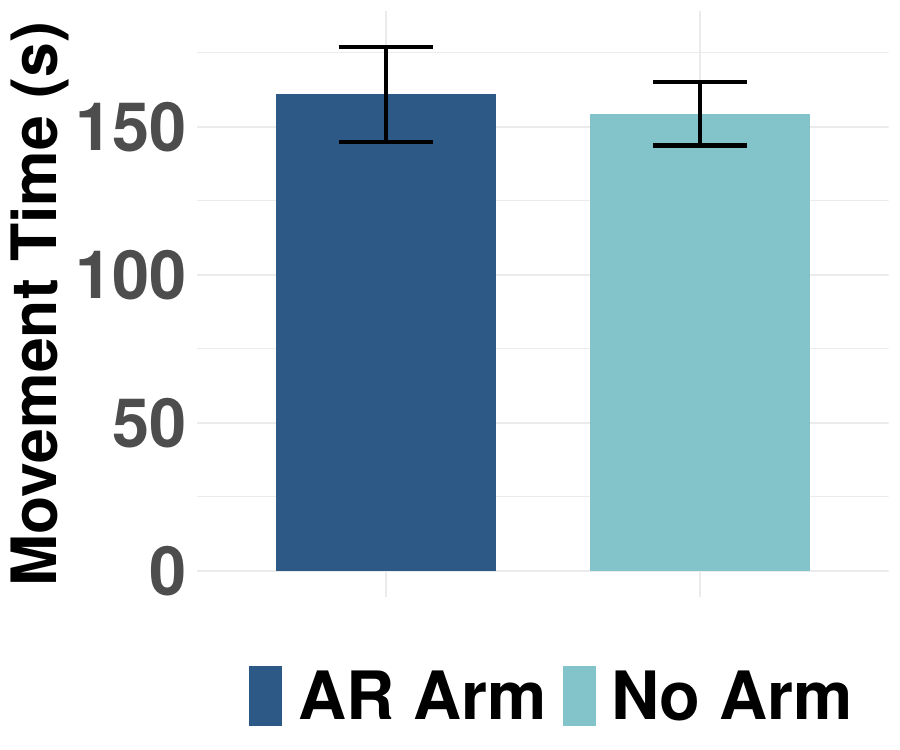}
    % \caption{}
    \label{fig:study2_mt_single}
\end{subfigure}
\begin{subfigure}[t]{0.52\linewidth}
    \centering
    \includegraphics[width=0.95\textwidth]{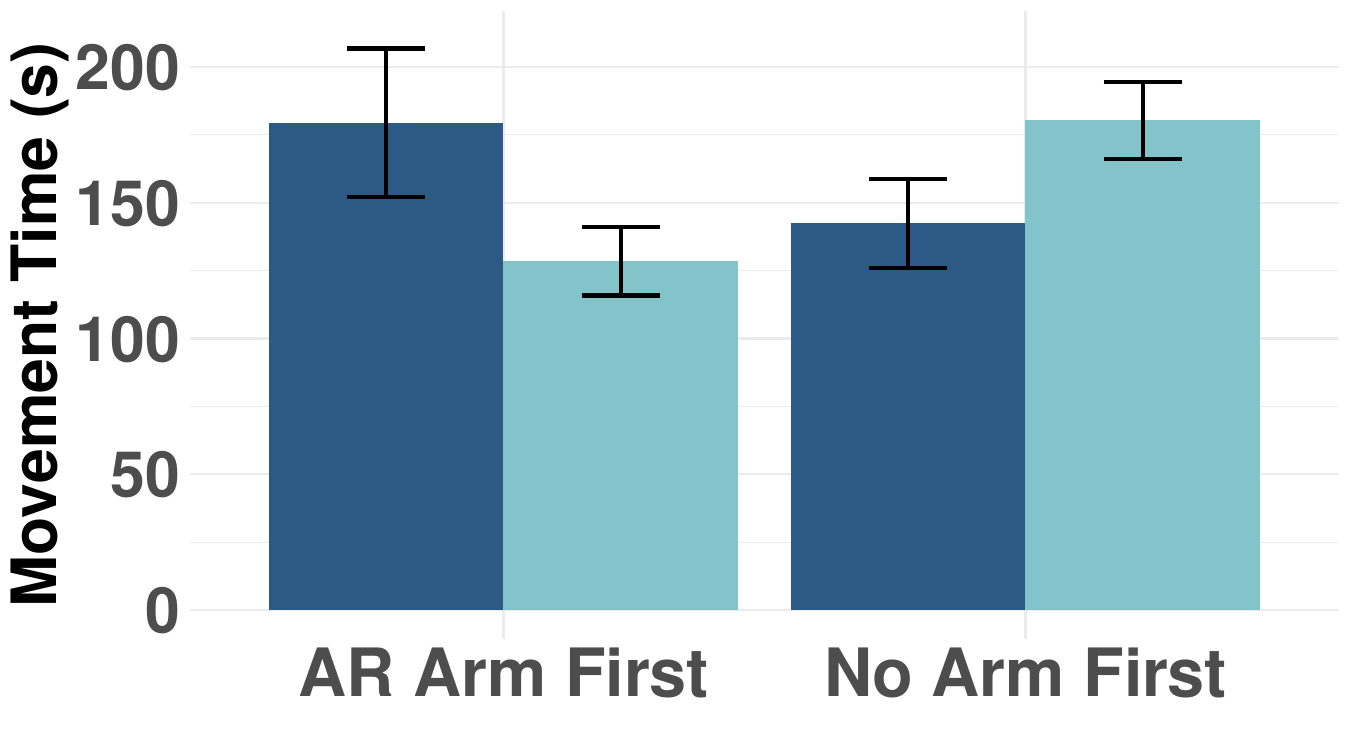}
    % \caption{}
    \label{fig:study2_mt_interact}
\end{subfigure}
\caption{Study 2 results: movement time by \textsc{Visualisation} (left), and separate plots by condition order (right).}
\vspace{-1.5em}
\label{fig:study2_mt}
\end{figure}

We found no statistically significant effects of \textit{Visualisation} or \textit{viz order} on \textit{Movement Time} (Figure \ref{fig:study2_mt}). The perceived \textsc{Physical Demand} was significantly affected by \textit{Visualisation} ($F_{1,23}=5.20, p<.05$), similarly for \textsc{Effort} ($F_{1,23}=5.06, p<.05$) and \textsc{Frustration} ($F_{1,23}=5.14, p<.05$).

\begin{figure}[t]
  \centering
  \includegraphics[width=\linewidth]{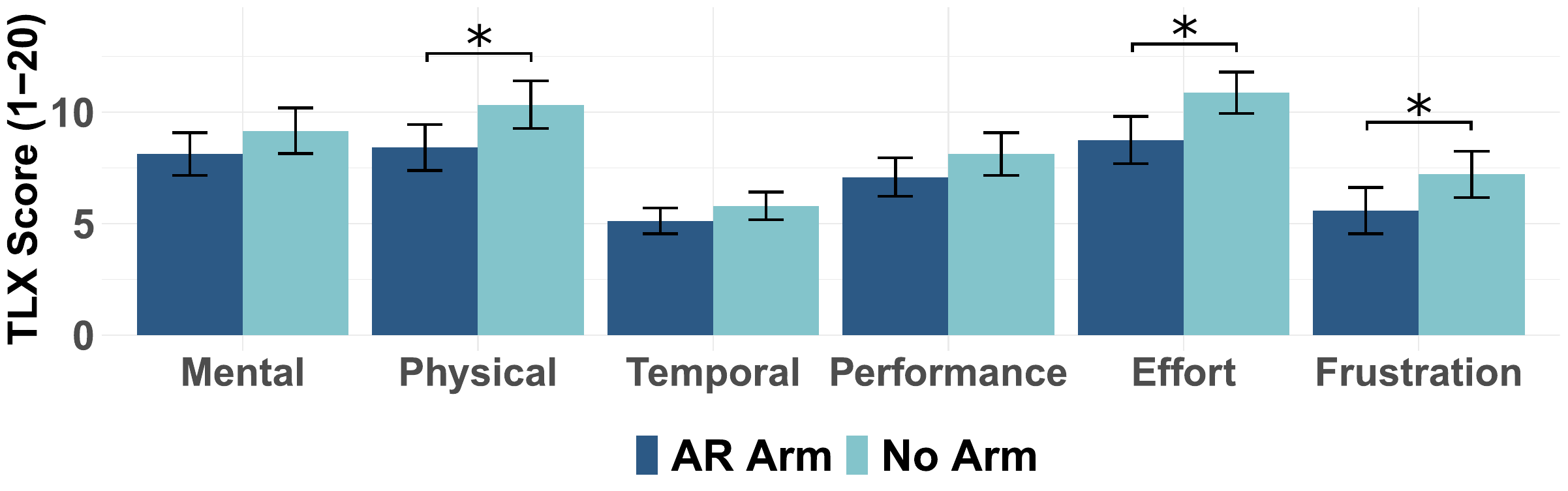}
  \caption{Study 2 results on the NASA-TLX questionnaire.}
  \vspace{-1.5em}
  \label{fig:study2_tlx}
\end{figure}

Thirteen participants found that the AR visualisation helpful ``\textit{It helped me with the position of the arm to have a reference. Without that, I couldn't imagine the best position} (P2).'' Six participants mentioned that the visualisation helped them learn the control at the beginning while the task becomes easy after a few rounds, and they did not need the visualisation any more. 

Fifteen participants found the AR visualisation helpful to learn the control mapping: ``\textit{It was very intuitive. I didn't have to learn anything beforehand, just having that reference was enough for the task} (P2).''; ``\textit{ Because the arm just looks like mine, so I can see if I was wrong} (P24).'' Six participants mentioned that AR helped them learn the control but only at the beginning: ``\textit{I think it's really important that you have that at the first time. But then you will know how it works} (P5).''

\subsection{Discussion of Study 2}

In Figure \ref{fig:study2_mt} (right), we can observe that while the earlier condition always yielded longer movement time, the difference seems more pronounced when \textsc{AR Arm} was administered first. This suggests that while individual differences and learning effects are present, experiencing the \textsc{AR Arm} first may have helped participants learn the control better. This observation is consistent with the subjective feedback. While 21 participants agreed that \textsc{AR Arm} was helpful for learning the control, six of them commented that it was only helpful at the beginning, and became redundant or distracting afterwards. These results suggests that the AR visualisation may serve as a useful learning aid for novice users to understand how the teleoperation control is done through the mappings between their arms and the robot, visually mediated by the \textsc{AR Arm}. 

The visual reference of the human-like appearance of the \textsc{AR Arm} made the task perceivably easier for participants, as supported by its significantly lower scores in \textsc{Physical Demand} and \textsc{Effort}. This is likely because they did not need to move their arms blindly to test the control at the beginning. They were able to quickly grasp how the rotations of their arm joints are mapped to the robot by observing the movement of the human-like \textsc{AR Arm}, which visualises the same structure as their own arms and is rendered on the physical robot. Similarly, they experienced less \textsc{Frustration} because the \textsc{AR Arm} provides straightforward visual feedback for their movements, saving the mental conversion effort.

%\section{Summary of Results from Study 1 and 2}

%From Study 1, we found that a virtual humanoid arm overlaying the physical robot is the most likely effective configuration for AR visualisation to help with MoCap-based teleoperation. Its evaluation in Study 2 indicated that the AR humanoid arm overlay on the robot arm improves user experience of MoCap-based teleoperation by providing an intuitive visual feedback that mediates the otherwise confusing mapping between the movement of two arms with different visual structures and in different orientations. Finally, observations from the performance data and subjective feedback from participants consistently suggest that the AR overlay of humanoid arm on the robot arm helps users learn its control at the beginning of teleoperation tasks, but may become redundant and no longer needed after users become familiar with the control.
%\section{Limitations and Future Work}

%The movement speed of the robot arm was limited by the refresh rate of the communication channel that sent MoCap data to the robot, and by the hardware limitation of the robot itself. While the results suggested that the AR visualization may be helpful for learning at early stages of the task, future work could specifically investigate the effectiveness of the AR arm to help learn MoCap-based teleoperation control by comparing transfers from different AR visualisations. 
\vspace{-0.1em}
\section{Conclusion}

Anthropomorphic robot arms suggests a novel approach of teleoperation through MoCap control that maps the rotations of the joints of the human operator's arms to a robot arm. We explore how AR can assist this by rendering a virtual arm as visual reference that mediates the inconsistencies between the human and the robot arm. In Study 1, we concluded that the optimal configuration of AR is a human-like arm overlaid on the physical robot in the same orientation. In Study 2, we evaluated this configuration and found that it helped reduce the perceived physical demand, effort, and frustration. Most participants found the AR arm suitable as a learning tool rather than an always-on visual guidance for teleoperation tasks. %Finally, we summarise key findings that directly inform future work that explore MoCap-based teleoperation of robot arms.

\interlinepenalty=10000
\balance
\bibliographystyle{ieeetr}
\bibliography{references}

\begin{thebibliography}{10}

\bibitem{pepper1984research}
R.~Pepper and J.~Hightower, ``Research issues in teleoperator systems,'' in {\em Proceedings of the Human Factors Society Annual Meeting}, vol.~28, pp.~803--807, SAGE Publications Sage CA: Los Angeles, CA, 1984.

\bibitem{Hedayati2018Improving}
H.~Hedayati, M.~Walker, and D.~Szafir, ``Improving collocated robot teleoperation with augmented reality,'' in {\em Proceedings of the 2018 ACM/IEEE International Conference on Human-Robot Interaction}, HRI '18, (New York, NY, USA), p.~78–86, Association for Computing Machinery, 2018.

\bibitem{gulletta2021human}
G.~Gulletta, E.~C.~e. Silva, W.~Erlhagen, R.~Meulenbroek, M.~F.~P. Costa, and E.~Bicho, ``A human-like upper-limb motion planner: Generating naturalistic movements for humanoid robots,'' {\em International Journal of Advanced Robotic Systems}, vol.~18, no.~2, p.~1729881421998585, 2021.

\bibitem{Hegel08Understanding}
F.~Hegel, S.~Krach, T.~Kircher, B.~Wrede, and G.~Sagerer, ``Understanding social robots: A user study on anthropomorphism,'' in {\em RO-MAN 2008 - The 17th IEEE International Symposium on Robot and Human Interactive Communication}, pp.~574--579, 2008.

\bibitem{zlotowski15anthropomorphism}
J.~Złotowski, D.~Proudfoot, K.~Yogeeswaran, and C.~Bartneck, ``Anthropomorphism: Opportunities and challenges in human–robot interaction,'' {\em International Journal of Social Robotics}, vol.~7, pp.~347--360, June 2015.

\bibitem{Roesler21meta}
E.~Roesler, D.~Manzey, and L.~Onnasch, ``A meta-analysis on the effectiveness of anthropomorphism in human-robot interaction,'' {\em Science Robotics}, vol.~6, no.~58, p.~eabj5425, 2021.

\bibitem{walker2023virtual}
M.~Walker, T.~Phung, T.~Chakraborti, T.~Williams, and D.~Szafir, ``Virtual, augmented, and mixed reality for human-robot interaction: A survey and virtual design element taxonomy,'' {\em ACM Transactions on Human-Robot Interaction}, vol.~12, no.~4, pp.~1--39, 2023.

\bibitem{Makhataeva20Augmented}
Z.~Makhataeva and H.~A. Varol, ``Augmented reality for robotics: A review,'' {\em Robotics}, vol.~9, no.~2, 2020.

\bibitem{Suzuki22Augmented}
R.~Suzuki, A.~Karim, T.~Xia, H.~Hedayati, and N.~Marquardt, ``Augmented reality and robotics: A survey and taxonomy for ar-enhanced human-robot interaction and robotic interfaces,'' in {\em Proceedings of the 2022 CHI Conference on Human Factors in Computing Systems}, CHI '22, (New York, NY, USA), Association for Computing Machinery, 2022.

\bibitem{yew2017immersive}
A.~Yew, S.~Ong, and A.~Nee, ``Immersive augmented reality environment for the teleoperation of maintenance robots,'' {\em Procedia Cirp}, vol.~61, pp.~305--310, 2017.

\bibitem{rosen2019communicating}
E.~Rosen, D.~Whitney, E.~Phillips, G.~Chien, J.~Tompkin, G.~Konidaris, and S.~Tellex, ``Communicating and controlling robot arm motion intent through mixed-reality head-mounted displays,'' {\em The International Journal of Robotics Research}, vol.~38, no.~12-13, pp.~1513--1526, 2019.

\bibitem{Gruenefeld20Mind}
U.~Gruenefeld, L.~Pr\"{a}del, J.~Illing, T.~Stratmann, S.~Drolshagen, and M.~Pfingsthorn, ``Mind the arm: realtime visualization of robot motion intent in head-mounted augmented reality,'' in {\em Proceedings of Mensch Und Computer 2020}, MuC '20, (New York, NY, USA), p.~259–266, Association for Computing Machinery, 2020.

\bibitem{Chen2021PinpointFly}
L.~Chen, K.~Takashima, K.~Fujita, and Y.~Kitamura, ``Pinpointfly: An egocentric position-control drone interface using mobile ar,'' in {\em Proceedings of the 2021 CHI Conference on Human Factors in Computing Systems}, CHI '21, (New York, NY, USA), Association for Computing Machinery, 2021.

\bibitem{OSTANIN18Interactive}
M.~Ostanin and A.~Klimchik, ``Interactive robot programing using mixed reality,'' {\em IFAC-PapersOnLine}, vol.~51, no.~22, pp.~50--55, 2018.
\newblock 12th IFAC Symposium on Robot Control SYROCO 2018.

\bibitem{suarez2020benchmarks}
A.~Suarez, V.~M. Vega, M.~Fernandez, G.~Heredia, and A.~Ollero, ``Benchmarks for aerial manipulation,'' {\em IEEE Robot. Autom. Lett.}, vol.~5, no.~2, pp.~2650--2657, 2020.

\bibitem{funk2021benchmarking}
N.~Funk, C.~Schaff, R.~Madan, T.~Yoneda, J.~U. De~Jesus, J.~Watson, E.~K. Gordon, F.~Widmaier, S.~Bauer, S.~S. Srinivasa, {\em et~al.}, ``Benchmarking structured policies and policy optimization for real-world dexterous object manipulation,'' {\em IEEE Robot. Autom. Lett.}, vol.~7, no.~1, pp.~478--485, 2021.

\bibitem{Yu2020Engaging}
D.~Yu, Q.~Zhou, B.~Tag, T.~Dingler, E.~Velloso, and J.~Goncalves, ``Engaging participants during selection studies in virtual reality,'' in {\em 2020 IEEE Conference on Virtual Reality and 3D User Interfaces (VR)}, pp.~500--509, 2020.

\bibitem{Zhou2024reflected}
Q.~Zhou, B.~V. Syiem, B.~Li, J.~Goncalves, and E.~Velloso, ``Reflected reality: Augmented reality through the mirror,'' {\em Proc. ACM Interact. Mob. Wearable Ubiquitous Technol.}, vol.~7, Jan. 2024.

\bibitem{NASA-TLX}
S.~G. Hart, ``Development of {NASA TLX}: Result of empirical and theoretical research,'' {\em California: San Jose State University}, 2006.

\end{thebibliography}

\end{document}